\documentclass[10pt,twocolumn,letterpaper]{article}

\usepackage{cvpr}
\usepackage{times}
\usepackage{epsfig}
\usepackage{graphicx}
\usepackage{amsmath}
\usepackage{amssymb}
\usepackage{subcaption}
\usepackage{booktabs}
\usepackage{array}
\usepackage{xspace}
\usepackage{cases}
\usepackage{multirow}
\usepackage{url}
\usepackage{xcolor}
\usepackage{paralist}
\usepackage[normalem]{ulem}
\usepackage{enumitem}
\newif\ifsubmit
\submitfalse

\ifsubmit
\newcommand{\sychien}[1]{}

\newcommand{\tsc}[1]{}

\newcommand{\cwwu}[1]{}

\newcommand{\ctliu}[1]{}

\newcommand{\wctu}[1]{}

\else
\newcommand{\sychien}[1]{{\bf \textcolor{purple}{S.-Y. Chien: #1}}}

\newcommand{\tsc}[1]{{\bf \textcolor{magenta}{T.-S. Chen: #1}}}

\newcommand{\cwwu}[1]{{\bf \textcolor{blue}{CW: #1}}}

\newcommand{\ctliu}[1]{{\bf \textcolor{brown}{C.-T. Liu: #1}}}

\newcommand{\wctu}[1]{{\bf \textcolor{red}{wctu: #1}}}

\fi

\newcommand{\mycaption}[2]{\caption{\textbf{#1.}~#2}}
\usepackage{pifont}

\newlength{\plotwidth}

\usepackage[breaklinks=true,bookmarks=false]{hyperref}

\cvprfinalcopy 

\def\cvprPaperID{3} 

\ifcvprfinal\fi
\begin{document}

\title{Viewpoint-aware Channel-wise Attentive Network for Vehicle Re-identification}

\author{
Tsai-Shien Chen\textsuperscript{1,2}, Man-Yu Lee\textsuperscript{1,2}, Chih-Ting Liu\textsuperscript{1,2}, Shao-Yi Chien\textsuperscript{1,2}\\
\textsuperscript{1}NTU IoX Center, National Taiwan University\\
\textsuperscript{2}Graduate Institute of Electronic Engineering, National Taiwan University\\
{\tt\small \{tschen, leemanyu, jackieliu\}@media.ee.ntu.edu.tw}\\
{\tt\small sychien@ntu.edu.tw }
}

\maketitle

\begin{abstract}
Vehicle re-identification (re-ID) matches images of the same vehicle across different cameras.
It is fundamentally challenging because the dramatically different appearance caused by different viewpoints would make the framework fail to match two vehicles of the same identity.
Most existing works solved the problem by extracting viewpoint-aware feature via spatial attention mechanism, which, yet, usually suffers from noisy generated attention map or otherwise requires expensive keypoint labels to improve the quality.
In this work, we propose Viewpoint-aware Channel-wise Attention Mechanism (VCAM) by observing the attention mechanism from a different aspect. Our VCAM enables the feature learning framework channel-wisely reweighing the importance of each feature maps according to the ``viewpoint'' of input vehicle.
Extensive experiments validate the effectiveness of the proposed method and show that we perform favorably against state-of-the-arts methods on the public VeRi-776 dataset and obtain promising results on the 2020 AI City Challenge.
We also conduct other experiments to demonstrate the interpretability of how our VCAM practically assists the learning framework.
\end{abstract}

\setlength{\plotwidth}{.85\linewidth}
\begin{figure}[t]
	\centering
    \includegraphics[width=\plotwidth]{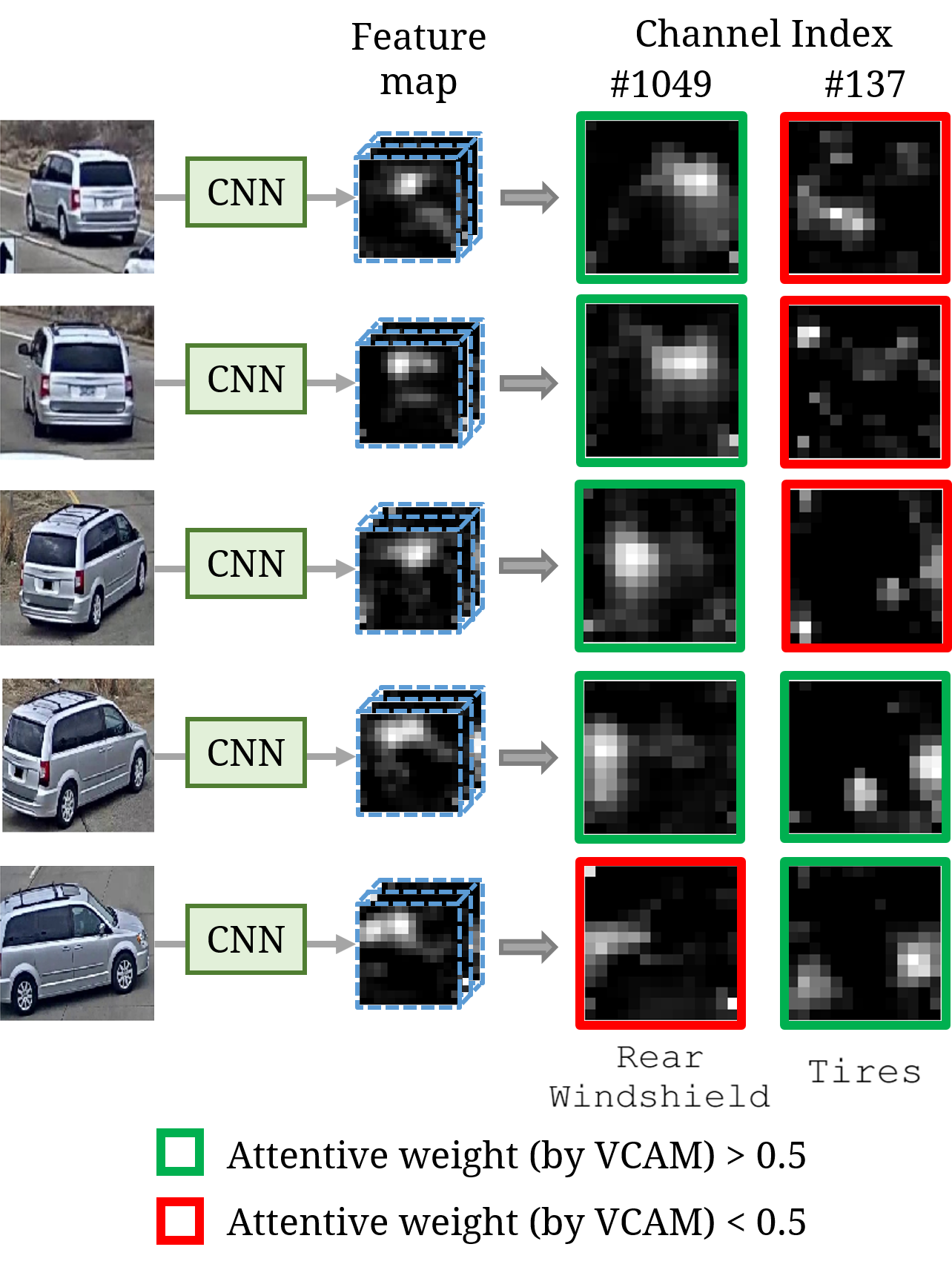}
    \mycaption{Illustration of Viewpoint-aware Channel-wise Attention Mechanism (VCAM)}{
    In the vehicle re-ID task, the channel-wise feature maps are essentially the detectors for specific vehicle parts, such as \texttt{Rear Windshield} and \texttt{Tires}. Our VCAM enables the framework to emphasize (i.e. attentive weight $>$ $0.5$) the features extracted from the clearly visible vehicle parts which are usually helpful for re-ID matching while ignore (i.e. attentive weight $<$ $0.5$) the others which are usually meaningless. 
    }
    \label{fig:concept}
\end{figure}

\section{Introduction}
\label{sub:intro}
Vehicle re-identification (re-ID) aims to match images of the same vehicle captured by a camera network. Recently, this task has drawn increasing attention because of its wide applications such as analyzing and predicting traffic flow.
While several existing works obtained great success with the aid of Convolutional Neural Network (CNN)~\cite{VeRi-776-1,VeRi-776-2,cityflow}, various challenges still hinder the performance of vehicle re-ID. One of them is that a vehicle captured from different viewpoints usually has dramatically different visual appearances.
To reduce this intra-class variation, some works~\cite{OIFE,AAVER,VAMI} guide the feature learning framework by
spatial attention mechanism to extract viewpoint-aware features on the meaningful spatial location. However, the underlying drawback is that the capability of the learned network usually suffers from noisy generated spatial attention maps. Moreover, the more powerful spatial attentive model may rely on expensive pixel-level annotations, such as vehicle keypoint labels, which are impractical in real-world scenario. In view of the above observations, we choose to explore another type of attention mechanism in our framework that is only related to high-level vehicle semantics.

Recently, a number of works adopt channel-wise attention mechanism~\cite{SENet,SCA-CNN,CBAM,RCAN} and achieve great success in several different tasks. 
Since a channel-wise feature map is essentially a detector of the corresponding semantic attributes, channel-wise attention can be viewed as the process of selecting semantic attributes which are meaningful or potentially helpful for achieving the goal. 
%
Such characteristic could be favorable in the task of vehicle re-ID.
Specifically, channel-wise feature maps usually represent the detectors of discriminative parts of vehicle, such as rear windshield or tires.
Considering that the vehicle parts are not always clearly visible in the image, with the aid of channel-wise attention mechanism, the framework should therefore learn to assign larger attentive weight and, consequently, emphasize on the channel-wise feature maps extracted from the visible parts in the image.
Nonetheless, the typical implementation of channel-wise attention mechanism~\cite{SENet,SCA-CNN} generates the attentive weight of each stage, explicitly each bottleneck, based on the representation extracted from that stage in the CNN backbone. 
We find that the lack of semantic information in the low-level representations extracted from the former stages may result in undesirable attentive weight, which would limit the performance in vehicle re-ID. 

As an alternative solution, in this paper, we propose a novel attentive mechanism, named \textit{Viewpoint-aware Channel-wise Attention Mechanism (VCAM)}, which adopts high-level information, the ``viewpoint'' of captured image, to generate the attentive weight.
The motivation is that the visibility of vehicle part usually depends on the viewpoint of the vehicle image.
As shown in Fig.~\ref{fig:concept}, with our VCAM, the framework successfully focuses on the clearly visible vehicle parts which are relatively beneficial to re-ID matching.
Combined with VCAM, our feature learning framework is as follows.
For every given image, our framework first estimates the viewpoint of input vehicle image.
Afterwards, based on the viewpoint information, VCAM accordingly generates the attentive weight of each channel of convolutional feature. Re-ID feature extraction module is then incorporated with the channel-wise attention mechanism to finally extract viewpoint-aware feature for re-ID matching.

Extensive experiments prove that our method outperforms state-of-the-arts on the large-scale vehicle re-ID benchmark: VeRi-776~\cite{VeRi-776-1,VeRi-776-2} and achieves promising results in the 2020 Nvidia AI City Challenge\footnote{https://www.aicitychallenge.org/}, which holds competition on the other large-scale benchmark, CityFlow-ReID~\cite{cityflow}. We additionally analyze the attentive weights generated by VCAM in interpretability study to explain how VCAM helps to solve re-ID problem in practice. We now highlight our contributions as follows:
\begin{compactitem}
\item
We propose a novel framework which can benefit from channel-wise attention mechanism and extract viewpoint-aware feature for vehicle re-ID matching.
\item
To the best of our knowledge, we are the first to show that viewpoint-aware channel-wise attention mechanism can obtain great improvement in the vehicle re-ID problem.
\item
Extensive experiments on public datasets increase the interpretability of our method and also demonstrate that the proposed framework performs favorably against state-of-the-art approaches.
\end{compactitem}

\setlength{\plotwidth}{0.95\textwidth}
\begin{figure*}[t!]
	\centering
    \includegraphics[width=\plotwidth]{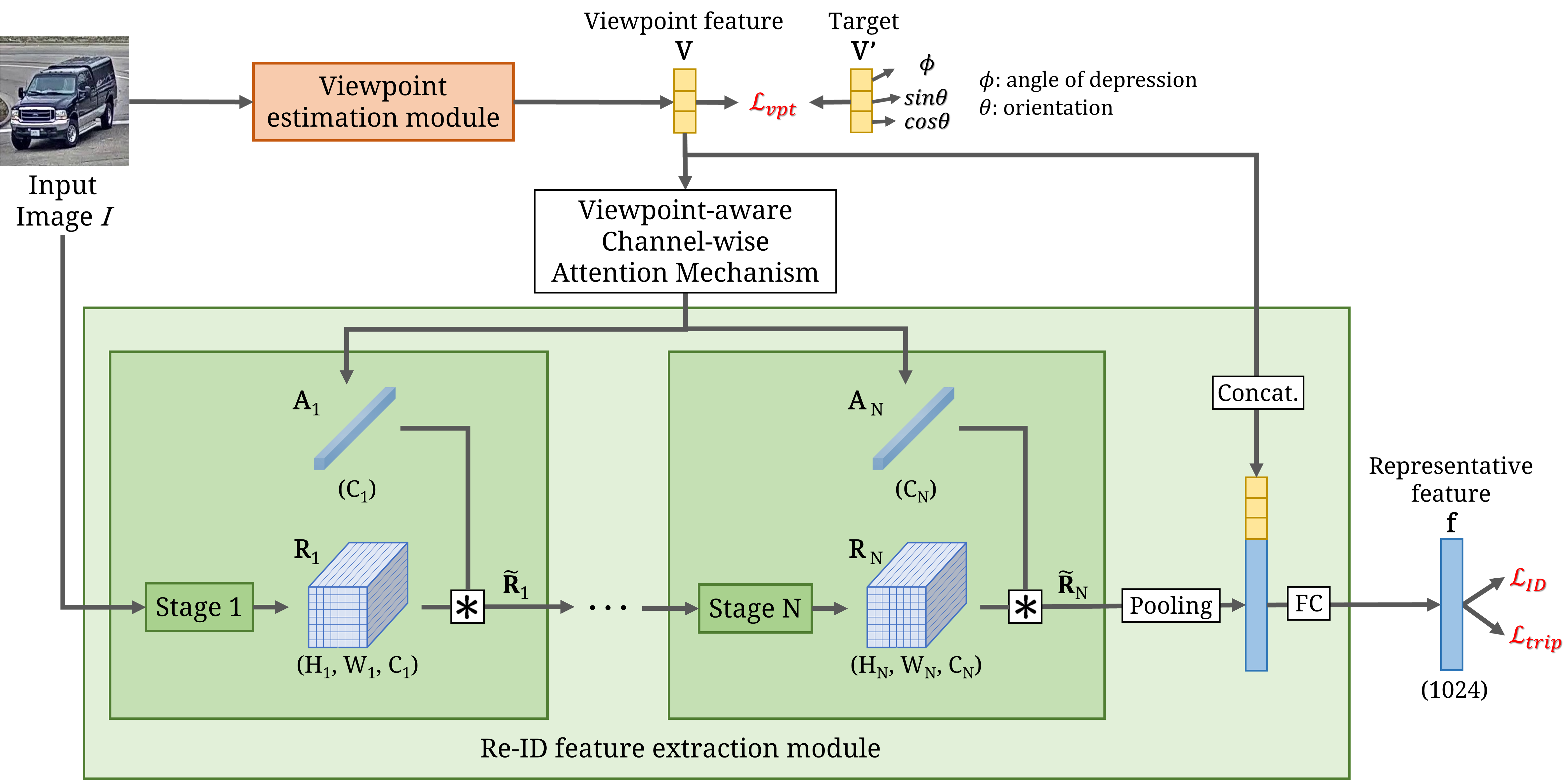}
    \mycaption{Architecture of our proposed framework}{}
    \label{fig:overview}
\end{figure*}

\section{Related Work}
\paragraph{Vehicle Re-Identification.}
Vehicle re-ID has received more attention for the past few year due to the releases of large-scale annotated vehicle re-ID datasets, such as VeRi-776~\cite{VeRi-776-1,VeRi-776-2} and CityFlow~\cite{cityflow} datasets.
As earlier work, Liu~\cite{VeRi-776-1}~\etal showed the advantage of using CNN model to tackle the vehicle re-ID problem.
However, vehicle captured from different viewpoint usually have dramatically different visual appearances which could impede the model capability of re-ID matching.

\paragraph{Viewpoint-aware Attention.}  
To reduce the impact caused by such intra-class variation, numerous works~\cite{OIFE,AAVER,VAMI,PAMTRI,split1,split2,split3} proposed the viewpoint-aware feature learning frameworks to adapt the viewpoint of input image.
Specifically, most of them utilized ``spatial'' attention mechanism to extract local features from the regions that are relatively salient.
For example, Wang~\etal~\cite{OIFE} and Khorramshahi~\etal~\cite{AAVER} generated spatial attentive maps for 20 vehicle keypoints to guide their networks to emphasize on the most discriminative vehicle parts.
While they are the first to show that viewpoint-aware features could aid vehicle re-ID, the required vehicle keypoint labels are expensive to obtain for real-world scenario.
To avoid such problem, Zhou~\etal~\cite{VAMI} proposed a weakly-supervised viewpoint-aware attention mechanism which can generate the spatial attention maps for five different viewpoints of vehicle. Instead of utilizing pixel-level annotations, they only requires image-level orientation information for training.
However, due to the lack of strong supervision on the generation of attention maps, the attention outcomes may become noisy and would affect network learning.
Considering to the general disadvantages of spatial attention mechanism mentioned above, we turn to a different aspect of attention mechanism to tackle the vehicle re-ID problem.

\paragraph{Channel-wise Attention.} 
Channel-wise attention can be treated as a mechanism to reassess the importance of each channel of the features maps. The benefits brought by such mechanism have been shown across a range of tasks, such as image classification~\cite{SENet}, image captioning~\cite{SCA-CNN}, object detection~\cite{CBAM} and image super-resolution~\cite{RCAN}.
Among existing works, typical implementation of channel-wise attention reweighs the channel-wise feature with the attentive weight which is generated by the representation extracted from each stage of CNN backbone.
However, as mentioned in Sec.\ref{sub:intro}, the lack of semantic information in the low-level representations extracted from the former stages may fail to generate meaningful attentive weight.
Accordingly, we exploit the high-level information, the ``viewpoint'' of image, to better assist the model to emphasize on those semantically important channel-wise feature maps.
\section{Proposed Method}
The whole feature learning framework is depicted as Fig.~\ref{fig:overview}.
For every given image $I$, there is a viewpoint estimation module to first evaluate the viewpoint of image and generate the viewpoint vector as $\textbf{V}$.
According to the information $\textbf{V}$, our viewpoint-aware channel-wise attention mechanism (VCAM) then generates the attentive weights of channel-wise feature maps extracted from each stage of re-ID feature extraction module.
Specifically, the CNN backbone of re-ID feature extraction module is constituted of $N$ stages, and the attentive weight $\textbf{A}_i \in \mathcal{R}^{C_i}$ generated by VCAM indicates the importance of channel-wise feature maps of the intermediate representation $\textbf{R}_i \in \mathcal{R}^{H_i \times W_i \times C_i}$ extracted from the $i^{th}$ stage in re-ID feature extraction module. 
Finally, the re-ID feature extraction module combined with the channel-wise mechanism would generate the representative feature $\textbf{f}$ for re-ID matching.
We will give more details about viewpoint estimation module in Sec.~\ref{pose}, viewpoint-aware channel-wise attention mechanism (VCAM) in Sec.~\ref{VCAM}, re-ID feature extraction module in Sec.~\ref{main}, and the overall training procedure of our framework in Sec.~\ref{learning}.

\subsection{Viewpoint Estimation Module}
\label{pose}
To better guide the VCAM generating the attentive weights of channel-wise feature maps with high-level semantic information, we utilize a viewpoint estimation module to embed the whole image into one representative viewpoint feature $\textbf{V}$ for every input image $I$. 
To confirm that the feature $\textbf{V}$ is able to explicitly indicate the viewpoint of image, we first define the target of viewpoint by two properties of captured vehicle image: angle of depression $\phi$ and orientation $\theta$. Angle of depression represents the angle between horizontal line and the line of camera sight. It can be easily obtained by the camera height $H$ and the horizontal distance between object and camera $D$ as:
\begin{equation}
\label{eq:depression}
    \mathcal{\phi} = arctan(H/D).
\end{equation}
Orientation indicates the rotation degree of the vehicle (from $0^o$ to $360^o$). 
However, we find that the discontinuity of orientation would seriously affect the learning of viewpoint estimation module. Specifically, for the image with orientation of $359^o$, the module would be mistakenly punished by huge loss when it predicts the orientation of $1^o$ even if there are only $2^o$ degree error between the real and predicted orientation. As a revised method, $sin\theta$ and $cos\theta$ are used to mutually represent the orientation which guarantee continuous differentiation for two similar viewpoints. Overall, the target of viewpoint feature is defined as:
\begin{equation}
\label{eq:viewpoint}
    \textbf{V}' = [\phi, sin\theta, cos\theta].
\end{equation}
With the target $\textbf{V}'$, we then apply the viewpoint loss:
\begin{equation}
\label{eq:viewpointloss}
    \mathcal{L}_{vpt} = \|\textbf{V} - \textbf{V}'\|_2,
\end{equation}
which represents the mean square error (MSE) between the prediction and target of viewpoint feature to
optimize our viewpoint estimation module.

\subsection{Viewpoint-aware Channel-wise Attention\\Mechanism (VCAM)}
\label{VCAM}
Based on the viewpoint feature $\textbf{V}$ extracted from the viewpoint estimation module, VCAM generates a set of attentive weights $\textbf{A} = [\textbf{A}_1, ..., \textbf{A}_N]$ to reassess the importance of each channel-wise feature map.
Compared to the typical implementation of channel-wise attention mechanism which uses the representations (extracted from the stages in CNN backbone) as reference to generate attentive weights,
our VCAM uses viewpoint information instead; the reason is that we expect our generated channel-wise attentive weight is positively related to the visibility of corresponding vehicle part, and, moreover, that part visibility is usually determined by the viewpoint of input vehicle image.
For example, in Fig.~\ref{fig:concept}, the attentive weight of the $137^{th}$ channel (which is the detector of tires) should be larger if side face of vehicle is clearly captured in the image.
All in all, according to the viewpoint feature $\textbf{V}$ with only three dimensions, our VCAM generates the attentive weights $\textbf{A}$ by a simple transfer function with one fully-connected (FC) layer:
\begin{equation}
\label{eq:weight}
    \textbf{A}_i = \sigma(\textbf{V}\cdot\textbf{W}_i),
\end{equation}
where $\textbf{W}_i \in \mathcal{R}^{3 \times C_i}$ denotes the parameters in FC layer and $\sigma$($\cdot$) refers to the sigmoid function.

\subsection{Re-ID Feature Extraction Module}
\label{main}
As shown in the Fig.~\ref{fig:overview}, the main purpose of the re-ID feature extraction module is to embed the final representation for re-ID matching with the aid of channel-wise attention mechanism.
Based on the viewpoint-aware attentive weights $\textbf{A}$ generated by VCAM, the module would refine the channel-wise features of the representations $[\textbf{R}_1, ..., \textbf{R}_N]$ extracted from the stages of re-ID feature extraction module.
Similar to previous works~\cite{SENet,SCA-CNN}, we use channel-wise multiplication between feature maps and attentive weights to implement channel-wise attention mechanism:
\begin{equation}
\label{eq:attention}
    \Tilde{\textbf{R}}_i = \textbf{R}_i \ast \textbf{A}_i,
\end{equation}
where $\ast$ represents convolution operator and $\Tilde{\textbf{R}}_i$ is the reweighted feature which would be fed into next CNN stage for further feature extraction.

After getting the feature extracted from the last stage, saying $\Tilde{\textbf{R}}_N$, the module first adopts adaptive pooling to suppress the feature.
To fully refer the viewpoint information, the feature is then concatenated with viewpoint feature $\textbf{V}$ and passed through one fully connected layer to get final representative feature $\textbf{f}$ used for re-ID matching.

\subsection{Model Learning Scheme}
\label{learning}
The learning scheme for our feature learning framework consists of two steps.
In the first step, we utilize large-scale synthetic vehicle image dataset released by Yao~\etal~\cite{synthetic} to optimize our viewpoint estimation module by the viewpoint loss ($\mathcal{L}_{vpt}$) defined in Eq.~\ref{eq:viewpointloss}:
\begin{equation}
    \mathcal{L}_{step1} = \mathcal{L}_{vpt}.
\end{equation}

In the second step, we jointly fine-tune the viewpoint estimation module and fully optimize the rest of our network, including VCAM and re-ID feature extraction module, on the target dataset with two common re-ID losses. 
The first one for metric learning is the triplet loss ($\mathcal{L}_{trip}$)~\cite{wtriplet}; the other loss for the discriminative learning is the identity classification loss ($\mathcal{L}_{ID}$)~\cite{discriminative}. 
The overall loss is computed as follows:
\begin{equation}
    \mathcal{L}_{step2} = \lambda_{trip}\mathcal{L}_{trip} + \lambda_{ID}\mathcal{L}_{ID}.
\end{equation}

\section{Experiments}

\subsection{Datasets and Evaluation Metrics}
\label{datasets}
Our framework is evaluated on two benchmarks, VeRi-776~\cite{VeRi-776-1, VeRi-776-2} and CityFlow-ReID~\cite{cityflow}.
VeRi-776 dataset contains 776 different vehicles captured, which is split into 576 vehicles with 37,778 images for training and 200 vehicles with 11,579 images for testing.
CityFlow-ReID is a subset of images sampled from the CityFlow dataset~\cite{cityflow} which also serves as the competition dataset for Track 2 of 2020 AI City Challenge. 
It consists of 36,935 images of 333 identities in the training set and 18,290 images of another 333 identities in the testing set. 
It has the largest scale of spatial coverage and number of cameras among all the existing vehicle re-ID datasets.

As in previous vehicle re-ID works, we employ the standard metrics, namely the cumulative matching curve (CMC) and the mean average precision (mAP)~\cite{scalable} to evaluate the results.
We report the rank-1 accuracy (R-1) in CMC and the mAP for the testing set in both datasets.
Note that in CityFlow-ReID dataset, the listing results are reported with rank list of size 100 on 50\% of the testing set displayed by the AI City Challenge Evaluation System.

\subsection{Implementation Details}
\label{details}
We respectively adopt ResNet-34~\cite{resnet} and ResNeXt-101 32x8d~\cite{resnext} as CNN backbone for the viewpoint estimation module and re-ID feature extraction module (both networks are pre-trained on ImageNet~\cite{imagenet} dataset). As for re-ID feature extraction module, we split the whole ResNeXt-101 into $N = 4$ stages; the sizes of representations extracted from each stage are $256 \times 56 \times 56$ (channel $\times$ height $\times$ width), $512 \times 28 \times 28$, $1024 \times 14 \times 14$, and $2048 \times 14 \times 14$ respectively.
Hence, the VCAM is composed by four independent networks which all take 3-dim viewpoint feature $\textbf{V}$ as input and generates a set of attentive weights $\textbf{A}$ with 256-dim, 512-dim, 1024-dim, and 2048-dim.

For training process of feature learning framework, we first optimize viewpoint estimation module with $\mathcal{L}_{step1}$ in advance on large-scale synthetic vehicle image dataset released by Yao~\etal~\cite{synthetic}, where viewpoint information is available.
Afterward, we optimize the rest of the framework, including VCAM and re-ID feature extraction module, and fine-tune the viewpoint estimation module (by 10 times smaller learning rate) with $\mathcal{L}_{step2}$ on target dataset.
For optimizing with triplet loss ($\mathcal{L}_{trip}$), we adopt the $PK$ training strategy~\cite{metric}, where we sample $P=8$ different vehicles and $K=4$ images for each vehicle in a batch of size $32$. 
In addition, for training identity classification loss ($\mathcal{L}_{ID}$), we adopt a BatchNorm~\cite{batchnorm} and a fully-connected layer to construct the classifier as in~\cite{bagoftricks,liu}.
We choose SGD optimizer with the initial learning rate starting from 0.005 and decay it by 10 times every 15000 iterations to train network for 40000 iterations.

\subsection{Ablation Study}
\label{ablation}
In this section, we conduct the experiments on both VeRi-776 and CityFlow-ReID datasets to assess the effectiveness of our Viewpoint-aware Channel-wise Attention Mechanism (VCAM) and show the results in Table~\ref{tab:ablation}.
We first simply train ResNeXt-101 without any attention mechanism as the baseline model and list the result in the first row. 
We also compare our VCAM with the typical implementation of channel-wise attention mechanism listed in the second row. For this experiment, the backbone is replaced with SE-ResNeXt-101~\cite{SENet} which shares similar network architecture with ResNeXt-101 except for adding extra SE-blocks, proposed by Hu~\etal\cite{SENet}, after each bottleneck block of ResNeXt-101.
It shows that compared to the baseline model, the performances are all boosted with the help of channel-wise attention mechanism.
However, while SE-ResNeXt-101 could only reach limited advancement ($1.7\%$ and $1.1\%$ for mAP on VeRi-776 and CityFlow-ReID), our proposed framework favorably achieves greater improvement on both datasets ($\mathbf{7.1\%}$ and $\mathbf{9.5\%}$ for mAP on VeRi-776 and CityFlow-ReID). It verifies that, according to the viewpoint information, our VCAM could generate more beneficial attentive weight to re-ID matching rather than the weight produced by typical channel-wise attention mechanism.

 \begin{table}[t]
    \centering
    \mycaption{Ablation study of our proposed VCAM ($\%$)}{}
    \label{tab:ablation}
    \begin{tabular}{l|cc|cc}
    \hline
    \multirow{2}{*}{Model} & \multicolumn{2}{c|}{VeRi-776} & \multicolumn{2}{c}{CityFlow-ReID} \\ \cline{2-5}
                   & mAP  & R-1  & mAP  & R-1  \\ \hline \hline
    ResNeXt-101    & 61.5 & 93.2 & 37.3 & 54.1 \\
    SE-ResNeXt-101 & 63.2 & 93.8 & 38.9 & 55.2 \\ \hline
    \textbf{VCAM (Ours)}    & \textbf{68.6} & \textbf{94.4} & \textbf{46.8} & \textbf{63.3}   \\
    \hline
    \end{tabular}
\end{table}

\begin{table}[t]
    \centering
    \mycaption{Comparison with state-of-the-arts re-ID methods on VeRi-776 ($\%$)}{
    Upper Group: attentive feature learning methods. Lower Group: the others. Note that all listed scores are from the methods \textbf{without} adopting spatial-temporal information~\cite{VeRi-776-2} and extra post processing such as re-ranking~\cite{rerank} .}
    \label{tab:sota}
    \begin{tabular}{l|ccc}
    \hline
    \multirow{2}{*}{Method} & \multicolumn{3}{c}{VeRi-776}\\ 
    \cline{2-4}
                            & mAP  & R-1  & R-5  \\ 
    \hline \hline
    OIFE~\cite{OIFE}        & 48.0 & 68.3 & 89.7 \\
    VAMI~\cite{VAMI}        & 50.1 & 77.0 & 90.8 \\
    RAM~\cite{RAM}          & 61.5 & 88.6 & 94.0 \\
    AAVER~\cite{AAVER}      & 61.2 & 89.0 & 94.7 \\
    GRF-GGL~\cite{GRF-GGL}  & 61.7 & 89.4 & 95.0 \\
    \hline
    GSTE~\cite{GSTE}        & 59.5 & -    & -    \\
    EALN~\cite{EALN}        & 57.4 & 84.4 & 94.1 \\
    MoV1+BS~\cite{baseline} & 67.6 & 90.2 & 96.4 \\
    MTML~\cite{MTML}        & 64.6 & 92.3 & 95.7 \\
    \hline
    \textbf{VCAM (Ours)}    &\textbf{68.6} & \textbf{94.4} & \textbf{96.9} \\
    \hline
    \end{tabular}
    \vspace{-3mm}
\end{table}

\setlength{\plotwidth}{1.\textwidth}
\begin{figure*}[t!]
	\centering
    \includegraphics[width=\plotwidth]{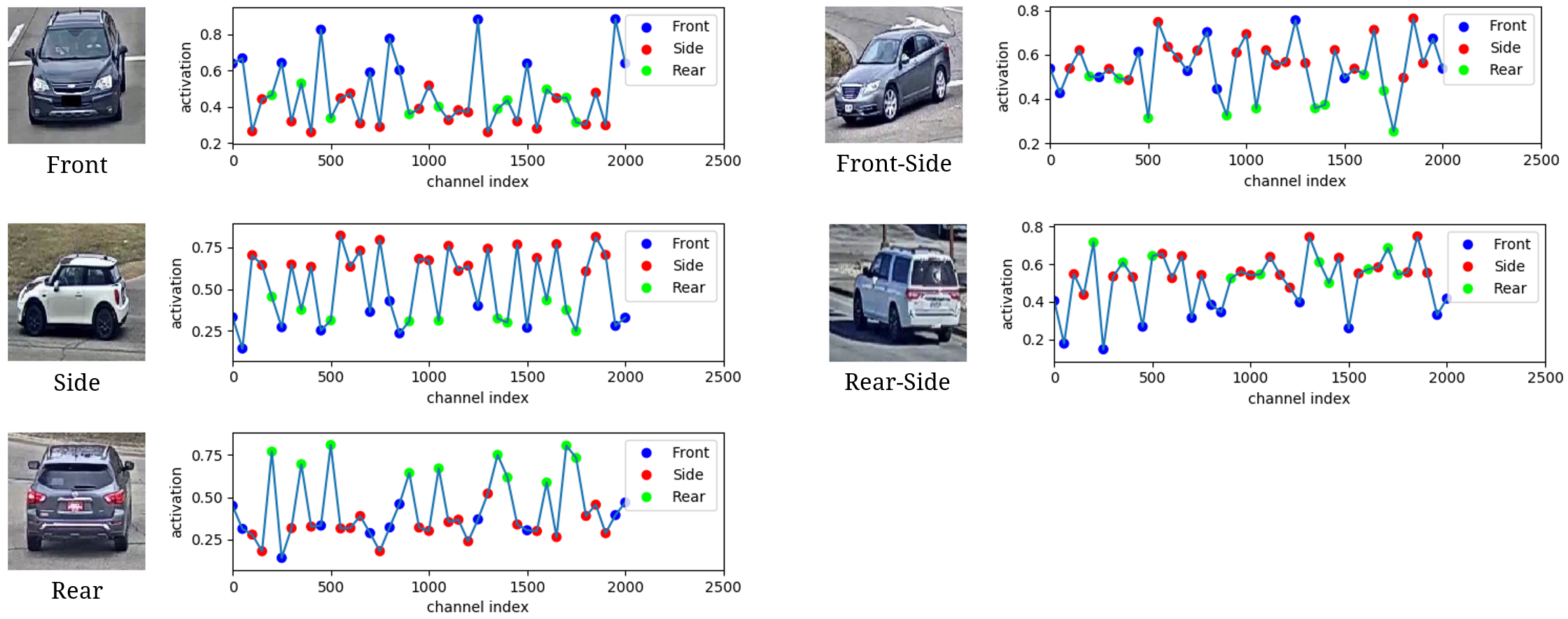}
    \mycaption{Distribution of Channel-wise Attentive weights}{
    We categorize vehicle images into five viewpoints, and, for each viewpoint, we sample 100 images and plot the average 2048-dim attentive weight vector for the fourth stage, namely $\textbf{A}_4$. 
    We assign and color each channel with one of front, side, or rear vehicle face label if the weight value of front, side, or rear viewpoint is relatively larger. We can then find that the channels emphasized by our proposed VCAM usually belong to the visible vehicle face(s).
    }
    \label{fig:visualization}
\end{figure*}

\subsection{Comparison with the State-of-the-Arts}
 \label{stoa}
We compare our method with existing state-of-the-art methods on VeRi-776 dataset in Table~\ref{tab:sota}.
Previous vehicle re-ID methods can be roughly summarized into three categories: attentive feature learning~\cite{OIFE,VAMI,RAM,AAVER,GRF-GGL}, distance metric learning~\cite{EALN,GSTE}, and multi-task learning~\cite{MTML}.
For the attentive feature learning which have been most extensive studied category, the existing methods all adopted ``spatial'' attention mechanism to guide the network to focus on the regional features which may be useful to distinguish two vehicles.
Nevertheless, unfavorable generated attention masks would hinder the re-ID performance on the benchmark.
In contrast, our proposed VCAM, which is the first to adopt channel-wise attention mechanism in the task of vehicle re-ID, achieves clear gain of $\mathbf{6.9\%}$ for mAP on VeRi-776 dataset compared to GRF-GGL~\cite{GRF-GGL} which is with attentive mechanism. It indicates that our framework can fully exploit the viewpoint information and favorably benefit from the channel-wise attention mechanism. 
Moreover, our proposed framework outperforms other state-of-the-art methods on VeRi-776 dataset.

\subsection{Interpretability Study and Visualization}
\label{visualization}
While our proposed framework have been empirically shown to improve the performance of vehicle re-ID task, we further conduct an experiment to illustrate how VCAM practically assists in solving the re-ID task in this section.
We first categorize the viewpoint into five classes: front, side, rear, front-side, and rear-side; example images of these five classes are shown in Fig.~\ref{fig:visualization}.
For each class, we then sample 100 images and compute the average 2048-dim attentive weight vector of the fourth stage, namely $\textbf{A}_4$. We uniformly select forty channels among total 2048-dim vector and plot the results in Fig.~\ref{fig:visualization}.
In order to increase the readability, we first analyze the attentive weights of three non-overlapped viewpoints, $\textbf{A}_4^{front}$, $\textbf{A}_4^{side}$, and $ \textbf{A}_4^{rear}$.
We assign and color each channel with one of front, side, or rear vehicle face label if the weight value of the corresponding viewpoint is relatively larger than the other two.
Take the $1^{st}$ channel shown in Fig.~\ref{fig:visualization} as example, it belongs to the front face and is, consequently, marked in blue because the attentive weight of front viewpoint is larger than the other ones of both side and rear viewpoints.
The physical meaning of the assignment of vehicle face label to each channel is that the channel-wise feature maps are essentially the detectors of vehicle parts, such as rear windshield and tires as illustrated in Fig.~\ref{fig:concept}, and, moreover, the visibility of that vehicle part is usually determined by whether the corresponding face is captured; for example, the presence of rear windshield in the image depends on whether the rear face is visible.
Hence, for each channel, we assign one of front, side, and rear vehicle face label.

With the assignment of vehicle face label, the following observation is made from the experiment result of all five viewpoints. For the attentive weight vector of each viewpoint, the relatively emphasized channels (commonly attentive weight values $> 0.5$) usually belong to the face(s) which can be seen in the image.
For example, for the images with front-side viewpoint, VCAM would generate larger attentive weight for the channels belonging to front or side face.
Based on the observation, we then accordingly give the explanation about the learning of our VCAM: our VCAM usually generates larger weights for the channel-wise feature maps extracted from clearly visible parts which are potentially beneficial to re-ID matching.

\subsection{Submission on the 2020 AI City Challenge}
\label{submission}
We also submit our proposed method to the 2020 AI City Challenge, which holds competition for vehicle Re-ID in the CityFlow-ReID dataset.
As a supplement to our proposed method, we employ some additional techniques for the final submission:
\paragraph{Synthetic dataset and Two-stage Learning}
Different from the past challenges held in previous years, the organizer release a large-scale synthetic vehicle re-ID dataset which consists of 192,151 images with 1,362 identities. All images on synthetic dataset are generated by an vehicle generation engine, called VehicleX, proposed by Yao~\etal~\cite{synthetic}, which enables user to edit the attributes, such as color and type of vehicle, illumination and viewpoint to generate a desired synthetic dataset.
With this engine, the attributes of synthetic images can be obtained easily without manually annotated which requires considerable or even prohibitive effort.
In this paper, we exploit viewpoint information of synthetic dataset to train viewpoint estimation module and identity information to enhance the learning of re-ID framework.
To better utilize the identity information of large-scale auxiliary dataset, which is synthetic dataset here, we adopt a two-stage learning strategy proposed by Zheng~\etal~\cite{vehiclenet} as our training scheme.
The framework is first trained with auxiliary dataset; when the learning converges, the classification FC layer used for training $\mathcal{L}_{ID}$ is replaced by a new one and the framework would be followingly trained with target dataset.
Based on the results displayed on the AI City Challenge Evaluation system, with the help of large-scale auxiliary dataset, we can achieve improvement of 5.3\% for mAP on the validation set of CityFlow-ReID (from 46.8\% to 52.1\%).

\paragraph{Track-based Feature Compression and Re-ranking}
Track-based feature compression is first proposed by Liu~\etal~\cite{manyu}.
It is an algorithm for the video-based inference scheme according to the additional tracking information of each image.
The whole algorithm includes two steps: merge and decompress.
First, all image features of the same track in the gallery would be merged into one summarized feature vector by average pooling to represent their video track. 
Then, in the decompression step, the summarized feature vector would be directly used as the representative feature for all images belonging to that video track.
With track-based feature compression, the rank list could be refined with the help of tracking information during inference scheme.
Finally, we perform typical re-ID scheme to rank the modified image features in the gallery according to the query image feature and adopt the k-reciprocal re-ranking method proposed by Zong~\etal~\cite{rerank} to re-rank our re-ID results.
Benefiting from track-based feature compression and re-ranking strategy, we can gain another improvement of 5.6\% for mAP on the validation set of CityFlow-ReID (from 52.1\% to 57.7\%).

Different from the listed results above, the score of our final submission to 2020 AI City Challenge Track2 is calculated with 100\% testing set. With our VCAM and the tricks mentioned above, we finally achieve $\mathbf{57.81\%}$ in mAP at the rank list size of 100 (rank100-mAP) and rank $15^{th}$ among all $41$ participated teams.
\section{Conclusion}
In this paper, we present a novel Viewpoint-aware Channel-wise Attention Mechanism (VCAM) which is the first to adopt channel-wise attention mechanism to solve the task of vehicle re-ID.
Our newly-design VCAM adequately leverage the viewpoint information of the input vehicle image and accordingly reassess the importance of each channel which is proven to be more beneficial to re-ID matching.
Extensive experiments are conducted to increase the interpretability of VCAM and also show that our proposed method performs favorably against existing vehicle re-ID works.

\section*{Acknowledgment}
This research was supported in part by the Ministry of Science and Technology of Taiwan (MOST 108-2633-E-002-001), National Taiwan University (NTU-108L104039), Intel Corporation, Delta Electronics and Compal Electronics.

{\small
\bibliographystyle{ieee_fullname}
\bibliography{egbib}
}

\end{document}